\documentclass[a4paper,12pt]{article}
\usepackage[margin=2.5cm]{geometry}
\usepackage{graphicx} 
\usepackage{setspace}
\usepackage{natbib}
\usepackage{times}
\usepackage{xcolor}
\usepackage{multicol}

\usepackage{linguex}

\title{The Human and the Mechanical: \\ \textit{logos}, truthfulness, and ChatGPT}
\author{Anastasia Giannakidou \& Alda Mari \\ University of Chicago \& Institut Jean Nicod CNRS/ENS/EHESS/PSL \\ Submitted to \textit{Intellectica}}

\begin{document}


\maketitle

\begin{abstract}
    The paper addresses the question of whether it is appropriate to talk about `mechanical minds' at all, and whether ChatGPT models can indeed be thought of as realizations of that. Our paper adds a semantic argument to the current debate. The act of human assertion requires the formation of a veridicality judgment. Modification of assertions with modals (\textit{John must be at home}) and the use of subjective elements (\textit{I hope that John is at home}) indicate that the speaker is manipulating her judgments and, in a cooperative context, intends her epistemic state to be transparent to the addressee. Veridicality judgments are formed  on the basis of two components: (i) evidence that relates to reality (exogenous evidence) and (ii) endogenous evidence, such as preferences and private beliefs. `Mechanical minds' lack these two components: (i) they do not relate to reality and (ii) do not have endogenous evidence. Therefore they lack the ability to form a belief about the world and a veridicality judgments altogether. They can only mimic that judgment, but the output is not ground in the very foundations for it.  

    \textbf{Keywords}: veridicality judgment, evidence, belief, truth
\end{abstract}




\section{Introduction: humans, machines, and \textit{logos}}

The promise of artificial intelligence (AI) since its inceptions in the 50s was to develop systems that are genuinely, i.e., human-level, intelligent. Second and third wave AI and machine learning came with this promise, but whether they delivered on it has been highly debated. In a recent book, \textit{The Promise of Artificial Intelligence,
Reckoning and Judgment} with MIT Press,  \cite{smith2019promise} makes the case that the artificial intelligence systems have failed to produce human-level intelligence and, he emphasizes, \textit{judgment}. Human intelligence, he argues, is of a different order than even the most powerful calculative ability of artificial computational systems.  Smith et al (\textit{ibid}.:p.XV) calls the AI ability \textit{reckoning} and argues  that it does not deliver human-quality judgment--- which is supposed to be  ``deliberative thought grounded in ethical commitment and responsible action.'' 

OpenAI's ChatGPT, Google's Bard and Microsoft's Sydney (collectively: ChatGPT) present the most recent achievements in the AI enterprise. They are large language models, that produce text and recognize patterns; they constitute, as \cite{chomsky2023noam} in a recent NYT Oped put it, \textit{marvels} of machine learning. Their function is to take enormous amounts of linguistic data, search for patterns, and eventually become proficient at generating statistically probable outputs that appear as intelligent and thoughtful text.  These programs, Chomsky et al. continue,  have been hailed as the ``first glimmers on the horizon of artificial general intelligence, that long-prophesied moment when \textit{mechanical minds} [emphasis ours] surpass human brains not only quantitatively in terms of processing speed and memory size but also qualitatively in terms of intellectual insight, artistic creativity and every other distinctively human faculty.'' 

In our paper, we want to address the question of whether it is appropriate to talk about `mechanical minds' at all, and whether ChatGPT models can indeed be thought of as realizations of that. What is a mechanical `mind'? Are the abilities of machines to perform calculations or generate probabilities enough to talk about the machines having a mind? Is the human mind simply a better version of the mechanical mind? These questions have a venerable history,  going back to Descartes's dualist thesis--- with the mind-body differentiation--- and further back  in Aristotle's theory of the mind (Greek: \textit{phyche}) which is a form of \textit{energeia} distinct from matter. Most language theorists and those working on language acquisition accept the existence of language as an argument of dualism, and the existence of the mind as a non-physical entity that, in humans but not in animals, encompasses language.  Animals have a mind too, but not one with \textit{logos}, argues Aristotle in a series of works.  \textit{Logos} is the feature of the human mind  responsible for both language and rational thought--- but also for the nature of humans as political animals,  as well as their ability to form moral judgments and distinguish right from wrong.  \textit{Logos} is, in other words, the conceptual causal prerequisite for human thinking, and moral and social flourishing, as can be seen very clearly in the passage below;  we retain the Greek word \textit{logos} which refers to both the ability to speak and the ability to think rationally:

\begin{quote}
And \textit{why the human [Greek: \textit{anthropos} emphasis ours]  is a political animal in a greater measure than any bee or any gregarious animal is clear}. For nature, as we declare, does nothing without purpose; and \textit{man alone of the animals possesses logos}. The mere voice, it is true, can indicate pain and pleasure, and therefore is possessed by the other animals as well, for their nature has been developed so far as to have sensations of what is painful and pleasant and to indicate those sensations to one another, but \textit{logos} is designed to indicate the advantageous and the harmful, and therefore also the right and the wrong; for \textit{it is the special property of the human in distinction from the other animals that he alone has perception of good and bad and right and wrong and the other moral qualities}, and it is partnership in these things that makes a household and a city-state. (Aristotle, Politics, Book I: 1253a).
\end{quote}

\textit{Logos} is the underlying principle of rationality that characterizes the thought of beings with language, i.e., human beings---  and it is this ability, according to Aristotle, that enables humans to form moral judgments of good and bad, and subsequently also able to apply these judgments to their societies (\textit{poleis}) in a way that serves the common good. \textit{Logos} thus encompasses language, the ability to calculate and draw logical conclusions, as well as the ability to form moral judgment. In Plato, likewise, \textit{logos} is the prerequisite for knowledge and wisdom (Greek: \textit{phronesis} which is best understood as instrumental and moral rationality, see Josiah Ober's recent discussion in \textit{The Greeks and the Rational} \citep{ober2022greeks}) is the ``only unqualified good'' (Meno 66-68). The human mind, therefore, because of \textit{logos},  is both a calculating and a judgment forming moral engine, and in addition it also has the capacity--- exclusive to animate beings--- to perceive, be aware, and self-reflect as we read in Aristotle's extensive discussion of perception in \textit{De Anima}. 

From this standpoint, the question of whether a non-animate machine `thinks' or `understands' receives an obvious answer: it does not, because it lacks \textit{logos}.  In what follows, we take the Aristotelian position as the point of departure and argue that it is only as a metaphor that we can talk about mechanical `minds', even in the most sophisticated incarnation of ChatGPT. AI can never by more than a calculating machine because by definition, being inanimate and lacking language, it  lack \textit{logos} and what follows from it, namely the ability to form any kind of judgment.  There are substantial qualitative differences between the human mind and the \textit{logos}-lacking data processing AI, among which the different \textit{teli} of the species: the \textit{telos} of humans is to exist in communities as \textit{logos}-driven beings who form judgments about what is true and false, and about what is right and wrong;  but the \textit{telos} of the machine is to process and produce statistically plausible outputs. 

Thinking, in other words, involves three abilities: to have language, to reason based on premises (data), and to form judgment, which includes judgment about truth, moral judgment, and self-reflection. Crucially, the human scale for empirical reasoning and judgment is typically small: we reason with a limited set of data, and often with gaps, as well as with non-rational premises (what \cite{GiannakidouMari2021know} call the \textit{affective component} of the epistemic basis for judgment, cf. \textit{infra}).   Again according to Aristotle and Plato, the human soul also contains the \textit{a-logon}, i.e., a non-rational component (of emotion, desires, etc.) which is not fully regulated by reason and can be in tension with it. This is an additional aspect in which the human mind is distinctive. 

John Searle, in his critique of the AI project to be discussed next, identifies what he calls the  `Strong AI' view \citep{searle1980minds,searle1989artificial,searle1990brain}. Strong AI is the position that suitably programmed computers can understand human  language (often called natural language), and that they have other mental capabilities similar to the humans whose behavior they mimic. According to Strong AI,  computers really do play chess intelligently, make decisions, or understand language. The Strong AI view views the human mind as merely a better version of the AI `mind', which at some point in the future and with more data and better programming, it will approximate more the human way of thinking.  There is a  weaker view, however--- we cann think of it as `weak AI'--- which says that AI is merely a useful tool for gathering and analyzing data because it  \textit{simulates} human abilities. In this view, we cannot simply transfer the  conclusions from AI to human cognition because they are qualitatively different; but we can extract information about language and reasoning that could or could not prove useful for the analysis of human cognition. This weaker position, we find, is indeed rational  and empirically justified--- as opposed to the strong AI hypothesis which is utopian and, we want to suggest,  irrational. 

In this brief note, we want to develop a novel kind of \textit{epistemological} argument that comes from the concept of truthfulness or, as we call it \textit{veridicality judgment}. Human communication and cognition is meaning-driven, as it follows from the Aristotelian \textit{logos} (as language, reason and judgment), and the basis for meaning is the veridical judgment, the evaluation of linguistic content as truthful or not.  The veridical judgment is referential to the world and relies on a process of proof, i.e., assessing evidence and intention of speakers as well as judging the evidence as reliable and trustworthy. If someone utters \textit{It is raining}, for instance, I will take it that she is uttering the sentence with the intent to inform and not mislead---- and I can always confirm whether the sentence is true or not by looking outside the window. Meaning is referential to the world, humans for beliefs about facts in the world, and they come to know \textit{with logos}, that is, with proof. We will argue, echoing \cite{smith1991}, that ChatGPT lacks this referentiality to the world, it can therefore not form beliefs  or come to know (see \cite{stalnaker1984}). It is epistemologically empty. 

Crucially, as we noted already, the process of forming a veridicality judgment relies on both declarative (i.e, knowledge or belief based thus objective) and affective (or emotive) premises which are subjective and can also be irrational or ignore evidence. The tension between the two can lead to formation of false beliefs (when people ignore evidence or favor unjustified but strong beliefs or dogma over evidence). ChatGPT, we will argue, is unable to form false beliefs either. 

The need to be truthful is the foundation for co-operative communication, argued Paul Grice in his classical piece \textit{Logic and Conversation} \citep{grice1975}; co-operativity says that interlocutors enter a conversation with the sincere intent of exchanging information, augment their knowledge space.  In our own work, we established the Veridicality Principle \citep{giannakidoumari2018b,giannakidoumari2021a}, as the foundation of co-operative conversation. The Veridicality Principle codifies conversation as a sequence of truth-based steps that matter also for action: to go back to the earlier example, if you ask me whether it is raining outside and I answer yes, then you assume that I am telling the truth, and you will have to make a decision based on that. You might decide not to go out, or if you might decide to go you must take an umbrella. For any co-operative conversation, interlocutors have to assume that they each are telling the truth, they have to rely on this basic truth-sharing intention, and the assertive content can always be verified of falsified by looking \textit{outside} the linguistic system, to the world.    Now, human beings are also able to deviate from co-operativity and in so doing  they can mislead or lie-- something that, we will argue, the sophisticated text producing AI is also unable to choose to do. If you lack the concept of truth (which needs the world), then you also lack the ability to know, to believe and to lie. 

 \section{Some recent history: can machines think?}

In modern analytical philosophy, the relation between the  mind (Aristotle's \textit{psyche}, Lat. \textit{anima}) and the body has been a major recurring theme, at least since Descartes' famous dictum \textit{cogito ergo sum} `I think therefore I am'. \textit{Cogito} is the incarnation of the Aristotelian \textit{logos}. Animate beings such as humans and animals seem different from non animate entities in  their abilities for self-driven thinking, action, and other types of  behavior. At different times, different features have been picked out as distinctive of human thinking, among which centrally the property of consciousness, self reflection and awareness.

 In 1950, the father of modern computing Alan Turing published a paper proposing a way of determining whether a computer thinks, which came to be known as  `The Turing test' \citep{turing2009computing}.  The idea is this: imagine a human being--- call it the judge--- engaged in conversation with two interlocutors hidden from view. One interlocutor is a human being, the other is a computer. The two enter the game, and the goal is for distinguish  which is which.
If a computer can fool 70 percent  of judges in a five-minute conversation into thinking it's a person, the computer passes the test. The question is: would passing the Turing test, which with the ChatGPT models seems within reach--- show that an AI has achieved thought and understanding?

John Searle suggested a negative answer  with his famous thought-experiment known as the Chinese Room Argument \citep{searle1989artificial}. The argument was first published in a 1980 article \textit{Minds, Brains and Programs} in the journal The Behavioral and Brain Sciences, and it has been one of the best-known arguments in modern  philosophy of language. Decades following its publication, the Chinese Room argument was the subject of many discussions, and in 1989, Searle included the argument in a book \textit{Minds, Brains and Science}. In January 1990, the popular periodical Scientific American took the debate to a general scientific audience, with an update contribution followed by a response \textit{Could a Machine Think?} written by philosophers Paul and Patricia Churchland\footnote{See  https://plato.stanford.edu/entries/chinese-room/  for more historical background on the exchange as well as specific objections and Searle's replies.}). 

The Chinese room argument is specifically addressed about knowledge of language, and is directed at what we mentioned earlier as the  `Strong AI' view, namely the position that suitably programmed computers can actually understand human  language and that in time they can reach human capabilities.  In the Chinese room argument, Searle imagines himself alone in a room following a symbol processing computer program (the analogue of Turing's `paper machine') for responding to Chinese characters that are slipped under the door. Searle himself does not know or understand Chinese, and yet, by following the program for manipulating symbols and numerals just like a computer does, he sends back appropriate strings of Chinese characters under the door, leading those outside to mistakenly suppose there is a person that knows Chinese in the room.

The point of the argument is that programming a digital computer may make it \textit{appear} to understand language while in fact it lacks knowledge and understanding of language. In other words, the `Turing Test' Searle argues, is inadequate as a method for correctly identifying  human thought because mimicking it can be quite convincing. Searle  assesses that computers have the ability to use syntactic rules to manipulate symbol strings--- but  human cognition is not simply syntactic computation, but must include understanding of meaning of the syntactic strings, in other words, it must include what we call in linguistics \textit{semantics}. Understanding of semantics inevitably involves reference to truth conditions and the world as proof and it is a property of the human mind; therefore the broader conclusion from the argument is that the strong AI position that equates human cognition to computation cannot be right. 

The Chinese language argument has had  large implications for semantics, the philosophy of language and modern theories of the mind and consciousness, and it is fair to say that it still stands strong as an argument for the position that human thought is more than computation--- which was, as we mentioned at the beginning obvious to Aristotle.  before we go on to develop our own argument about semantics and the formation of the veridicality judgment,  we will  consider in more detail the argument from language acquisition developed in the the modern linguistic tradition. This argument specifies the additional qualities of the human mind as \textit{innate} principles guiding a particular type of learning for humans that is radically different-- in both manner and scope-- from the modes of processing of advanced AI. 

\section{The argument from  learning language: innateness and recollection}

Noam Chomsky, Steven Pinker,\footnote{e.g. \cite{pinker2003mind}}  and most current linguistic theory posit \textit{innate} properties in the human mind, such as the ability to learn language, that  guide learning under exposure to very limited set of data. This is known as the poverty of the stimulus argument, and it connects well to what Chomsky himself call's `Plato's problem', to be considered here. 

The innateness hypothesis emerged in great part as a response to the behaviorist model of language learning which claims that language learning is imitation based on positive reinforcement (reward of desirable behavior and punishment of undesirable one). The behaviorist claim is, roughly, that  the child imitates the language of its parents or carers, and that successful attempts are
rewarded because an adult who recognizes a word spoken by a child will praise the child and/or give it
what it is asking for (\cite{skinner}). Successful utterances are therefore reinforced while unsuccessful ones are ignored.  In the behaviorist model, in other words, language acquisition is a form of training, a kind of instruction. Chomsky, Pinker and an explosion of linguists working on language acquisition since the late 50s established a number of arguments for why this is not the correct characterization of language learning. The most notable argument is that children are not exposed to a lot of data to begin with,   but are still able to learn quickly, in predictable patterns, and with rapid generalizations, manifest  in errors made by children--- which reveal that the children are actively figuring out rules.  For instance, a child may say `growed' instead of the correct `grew' not because they are copying an adult (who would never utter it), but rather because they over-apply a rule. 

Secondly, the work on language acquisition has revealed that the vast majority of children go through the same developmental stages of acquisition. We refer to these as developmental milestones,  sequential phases such as word learning, building two word combinations, acquisition of articles, inflections, etc. Children go through the same  stages at specific age frames in roughly every language we consider, and bilingual language acquisition follows the same path. If language learning is imitation and training, we could not expect such universal patterns; rather we would expect acquisition to be quite variable, affected by cultural and behavioral patterns in the environments that depend on  the
treatment the child receives or the customs of the society in which the child grows up.

Thirdly, children are in fact often unable to repeat what an adult says, especially if the adult utterance
contains a structure the child has not yet acquired. Well known cases involve the following which concerns acquisition of negation:

\ex. Child: Nobody don't like me\\
Mother: No, say, "Nobody likes me."\\
Child: Nobody don't like me.\\
(Eight repetitions of this dialogue)\\
Mother: No, now listen carefully: say, "Nobody likes me."\\
Child: Oh! Nobody don't likes me. (data reported in \cite{mcneill1968development})

The child here not only does not imitate, but it is unable to see what the error is. Very few children are receptive to language instruction in early acquisition, again demonstrating that language acquisition is not a process of instruction but rather self-driven \textit{maturation}. To this end, there has  also been important evidence for a critical period for language acquisition. Children who miss milestones for whatever reason (e.g. such as Genie discovered in 1970 at the age of 13 having been deprived of linguistic exposure), would never catch up. 

Noam Chomsky  famously reviewed  Skinner's book \citep{chomsky1980review}). He  focused particularly on the poverty of the stimulus argument, and emphasized the fact that adults do not typically speak in grammatically complete sentences.  Chomsky and Pinker independently conclude that children must have an inborn, innate faculty for language, a language instinct as is the title of Pinker's book \cite{pinkerinstinct}.  The child's natural predisposition to learn
language is triggered by linguistic input in the environment, and the child's mind is able to learn and generalize and reach adult linguistic competence with very little data and without instruction. It is simply false, says Chomsky, that `careful arrangement of contingencies of reinforcement by the verbal community is a necessary condition of language learning.' \cite{chomsky1980review}, p. 39. Children learning language do not appear to be being `conditioned' at all, and explicit training, as we saw, is simply not effective. 

Noam Chomsky himself gave the name `Plato's Problem' to the general problem of learning  in conditions of impoverished data, echoing Plato's theory of learning known as  the \textit{recollection} (anamnesis) theory. The recollection theory is developed in two dialogues, Meno and Phaedo, and says that learning is possible not as a discovery, but as recall of previous knowledge. Putting aside the particular framing of Plato's theory of Forms (for which recollection is an argument), in Phaedo (72e- 78b) Socrates argues that  ``we must at some previous time have learnt what we now recollect'', and this is possible ``only if our soul existed somewhere before it took on the human shape'' (72d). In this passage, the soul is compared to Forms  which are eternal and also exist \textit{prior to birth} (\textit{pro tou genesthai} in the Greek text), and do not have physical form (thus being distinguished from the particulars which have physical form; recall mind versus body). The recollection argument is the innateness argument: it says that when one knows something they do so because they recall from knowledge before they were born, i.e., knowledge that is inherent in their mind.  This nativist position is later taken by Kant who in his \textit{Critique of Pure Reason} argues that what guides our learning is a number of pre-existing cognitive categories.  Putting further details aside,  the basic point here is that the innateness hypothesis of language entails that the human mind is self-driven for language learning by innate principles that are not themselves physical data, and it is not blind processing large set of data and drawing statistical conclusions. 


\begin{quote}
ChatGPT and similar programs are, by design, unlimited in what they can ``learn'' (which is to say, memorize); they are incapable of distinguishing the possible from the impossible. Unlike humans, for example, who are endowed with a \textit{universal grammar that limits the languages we can learn} to those with a certain kind of almost mathematical elegance, these programs learn humanly possible and humanly impossible languages with equal facility. Whereas humans are limited in the kinds of explanations we can rationally conjecture, machine learning systems can learn both that the earth is flat and that the earth is round. They trade merely in probabilities that change over time.
\end{quote}

 In this passage we see the reinforcement of the nativist position: in comparison to the ChatGPT whose learning is exclusively empirical, the human mind is endowed with guiding learning principles (universal grammar) which enable both learning, as well as judgment of what is a possible grammar and what is not.  The  crux of machine learning, they continue "is description and prediction; it does not posit any causal mechanisms or physical laws. Of course, any human-style explanation is not necessarily correct; we are fallible. But this is part of what it means to think: To be right, it must be possible to be wrong. Intelligence consists not only of creative conjectures but also of creative criticism. Human-style thought is based on possible explanations and error correction, a process that gradually limits what possibilities can be rationally considered."  Again the possibility of error arises as evidence for the innate ability to form judgment, which is a property of the human mind but not of ChatGPT. 

With this background, we will now proceed to discuss the formation of the veridicality judgment which is a basic ability to distinguish truth from falsity.

\section{The veridicality judgment: truth, evidential and subjective components}

The arguments that we discussed all center on showing that the way way human cognition works is different from the way data processing AI works, they are arguments based on the \textit{modes of function}: the AI merely processes syntactic rules and establishes statistical probabilities, and it works with large sets of data---  while the human mind works with small sets of data and guided by innate principles.  Both objections are about the \textit{modus operandi}. 

We will now develop a novel kind of epistemological and conceptual argument, as we promised at the beginning, that to some degree echoes Cantwell-Smith's objection that AI lacks the concept of truth. We will develop this argument by further arguing that the truth judgment (veridicality) relies crucially on two components--- a rational and a potentially irrational one or, as we phrased it in \cite{mari2016assertability,GiannakidouMari2021know}, an exogenous  and an endogenous component. The former is the objective basis for truth and assertoric force, and it is evidential: it contains what speakers know and believe with mind-independent evidence --- while  the  latter contains subjective preferences of speakers, their desires, expectations, and also preferences based on ideology, tastes, non-factual beliefs,  and the like.  The evidential component is referential to the world and relies on a process of proof, i.e., assessing evidence and forming a conclusion. The veridicality judgment consists in both forming conclusions based on evidence, but also judging reliability of content, source of information, and ultimately establishing trustworthiness based on subjective preferences of the endogenous component. 

  Aristotle gives a well-known definition of truth in his \textit{Metaphysics} (1011b25):  ``To say of what is that it is not, or of what is not that it is, is false, while to say of what is that it is, and of what is not that it is not, is true.'' Very similar formulations can be found in Plato (Cratylus 385b2, Sophist 263b). The Aristotelian truth serves as the foundation for the main paradigm of \textit{truth-conditional semantics} that we use in linguistic analysis, and Tarksi's \textit{correspondence theory} of truth. Truth, in the correspondence theory,  consists in a direct relation of a sentence to reality: the sentence \textit{Snow is white} is true if and only if the snow in the world is white. This well-motivated understanding is central to natural language semantics, and implies metaphysical realism that acknowledges \textit{objective} truth. \footnote{Objective truth correlates with \textit{fact} but also with \textit{time}: simple positive present and past sentences such as \textit{Ariadne arrived in Paris last night, Ariadne is eating breakfast right now},  are true or false objectively, which means that the sentences, if true, denote facts of the world. Future sentences, on the other hand, such as \textit{Ariadne will go to Paris next week} are objectively false at the time of utterance (since they have not happened yet), but could or must be true---depending on the  strength of prediction---at a future time.}
  
  In linguistic semantics and pragmatics we assume that the assertion of a bare sentence requires that the speaker follows the Gricean principle of Quality \citep{grice1975},  which is a fundamental principle for co-operative conversation. Quality demands that the speaker be truthful, and this means that when uttering a bare sentence, she knows, or has grounds to believe that it is raining or that it rained. In \cite{giannakidoumari2021a,GiannakidouMari2021know} we argued that this is the prerequisite for assertability:

\ex. Veridical commitment as a prerequisite for assertion:\\
A sentence $S$ is assertable if and only if the speaker is \textit{veridically committed}  to the proposition  $p$ denoted by $S$.

By uttering the sentence ``It is raining'', the speaker is veridicality committed to the truth of the content \textit{It is raining} if she has evidence that supports it (see \cite{giannakidou1998,giannakidou2013,giannakidoumari2018b,giannakidoumari2018unified,giannakidoumari2021a}); if the speaker doesn't know that it is raining but utters the sentence, she is lying or misleading, which means she has the intention to deceive.\footnote{For an overview of the vast literature, see \cite{garcia2023lying}.} Whether ChatGPT has a self-driven ability to do that seems highly questionable, and we briefly address this question later in the discussion.

When the speaker is veridically committed she is also assertorically committed, and can think of the veridical commitment as the mental state (or, attitude) of commitment to truth that drives the force of assertion. On the other hand,  when a speaker chooses to modalize she indicates an epistemic or doxastic state that cannot veridically commit, i.e, she is taking a \textit{nonveridical stance}. She is now uncertain about whether it is raining or not. This epistemic uncertainty is gradient: with \textit{may} or \textit{might} or an expression such as \textit{It is possible that it rains},  raining is considered a mere possibility, and the commitment is trivial in the sense that the possibility is not excluded. But when a necessity modal is used, raining  is considered very likely, and with the future modal it is to be expected. In this case, we have talked about \textit{bias}: the linguistic anchor is biased towards $p$.
Consider the following declarative sentences including two with the modal verbs (\textit{must} and \textit{may}), and one with the future modal
 \textit{will}:

\ex. \a. It is raining.
\b. It rained.
\b. It must be raining.
\c. It may be raining. 
\d. It will rain.

Let us call the tensed sentences without modal verbs  `bare'.\footnote{For an in depth discussion of modality in this framework, see \cite{giannakidoumari2021a}.}
The assertoric force with modals is to a weaker proposition, as is obvious. Epistemic modal verbs are indicators that the speaker reasons with uncertainty and that she leaves open both options, $p$ and not $p$, as indicated in the following examples \citep{lassiter2016must}:

\ex. \a.This is a very early, very correct Mustang that has been in a private collection for a long time. . . . The speedo[meter] shows 38,000 miles and \textit{it must be 138,000, but I don't know for sure}.
\b. \textit{I don't know for sure}, sweetie, but she \textit{must have been} very depressed. A person doesn't do something like that lightly.
\c. It must have been a Tuesday (but I don't know for sure), I can't remember.

The veridical commitment to truth is motivated by information that the speaker possesses, what we have called  the \textit{body of evidence}. In the case of \textit{must},  veridical  bias is grounded on reasons for \textit{preferring} that the proposition be true, but it is not knowledge of it being true. Modals are, as we argued anti-knowledge markers; a speaker chooses to use an epistemic modal because the body of evidence does not allow full veridical commitment to $p$. In support of this, consider also the following case, which has been discussed in the literature quite a lot (see \cite{karttunen1979,von2010must}. Direct visual perception contexts are famously cited as rejecting modalization:
 
 \ex. \label{directrain} Context: $i$ is standing in front of the window and sees the rain.
\a. \#It must be raining. 
\b. \#It may be raining. 
\c. \#It might be raining.

The modals are infelicitous here because if I see the rain, I \textit{know} that it is raining, and knowledge creates veridical commitment; modalization is prohibited because direct evidence is a reliable path to knowledge.  

Commitment and bias allow us  to define truthfulness as a scale, and we summarize the gist below. Veridical commitment is epistemic or doxastic commitment: the speaker knows $p$ or believes it to be true, she is in a veridical state and therefore \textit{fully} committed to $p$. This is the strongest form of commitment; then we have veridical bias and non-veridical equilibrium which is the weakest \textit{trivial} commitment, i.e. not excluding $p$:

\ex. Scale of veridical commitment.(\cite{giannakidoumari2016cMUSTFUT}): \\
	$<$$p$, MUST $p$, MIGHT $p$$>$;\\
	 where $i$ is the speaker, $p$ conveys full commitment of $i$ to $p$; MUST $p$ conveys \textit{partial} commitment of $i$ to $p$, and MIGHT $p$ conveys \textit{trivial} commitment of $i$ to $p$.
	 
	\ex. Nonveridical equilibrium. \\
$p$ and its negation are open possibilities, and \textit{there is no bias, i.e. the two options are considered equal possibilities.}

The key here is the equalitarian clause that the two options are considered equal possibilities; there are no prior beliefs or expectations that might turn the odds in favor of the positive or negative possibility.  Nonveridical equilibrium characterizes also non-assertive sentences such as questions: \textit{Did it rain yesterday?, I wonder whether it rained yesterday}. In the nonveridical stance,  the mental state allows both options $p$ (it rained) and $\neg p$ (it did not rain). In the case of possibility modals, there is no bias towards $p$, we have a state or `balanced uncertainty'.

Veridical commitment is a judgment, i.e.,  an evaluation of what is perceived or understood as true by a linguistic agent. It can be influenced, as we will now argue,  by \textit{endogenous} factors and biases that can create purely solipsistic (that is, without evidence) and potentially false beliefs.

As we argued in \cite{GiannakidouMari2021know}, the body of evidence contains two types of content. First, there is declarative informative content, namely what the speaker or subject of attitude knows to be true, believes to be true, remembers, or  understands to be true--- what kind of information they have, in other words. If commitment to truth is sincere, this type of content is evidential. The declarative component  is \textit{rational}, hence it contains logical deductive rules, as well as inductive, stereotypical generalizations that guide rational thought. For instance, if the speaker has heard that Ariadne read War and Peace from a reliable source, they will take this hearsay information to be true and convey it as true; but if they hear the same sentence uttered by a pathological liar, rationality should make the speaker reluctant to commit themselves to the sentence. Likewise, if I wake up in the morning and I see the streets being wet, I can truthfully report this by saying \textit{It rained last night}, because it is rational to infer, by the wetness of the street, that it rained--- and less rational to say, for example, that \textit{it snowed}, or \textit{The neighbors left the water running again}.

Hence linguistic agents form veridicality judgments based on information they have and general rules of inference, and chose accordingly to commit fully, partly or trivially to a proposition. But there is another type of content that is relevant and may interfere with the body of evidence. This is subjective, and broadly understood as  \textit{emotive} content bringing in the speaker's internal \textit{perspective}.  This content is highly subjective and to a large extent \textit{private}: it contains the set of desires and hopes of the linguistic agent, as well as their political, ideological, religious, or aesthetic beliefs, including prejudice and biases. This component is not so relevant to whether Ariadne read War and Peace, or to whether John will be here by 5 pm, and plays no role in the assertability veridicality condition in these innocuous cases. "If I hate Russian writers, and I know that Ariadne read War and Peace but I wish she hadn't, I still utter \textit{Ariadne read War and Peace} truthfully" \cite{GiannakidouMari2021know}. 

But the emotive component becomes relevant when the sentence contains predicates that denote subjective evaluation:

\ex. War and Peace is a masterpiece. \label{masteryes}

\ex. The duck is delicious.

\ex. Hamas's attack on October 7, 2023 was an act of resistance / an act of terror. 

These sentences express opinions. Opinions  do not depend on evidence alone; rather are highly dependent on the subjective emotive component which need not be rational. For instance, if I irrationally hate Russian writers, most likely  I will not utter \ref{masteryes},  and upon hearing it I will object with \ref{masterno} below. 

\ex. War and Peace is not a masterpiece. \label{masterno}

Likewise, if I am a vegetarian I will disagree with the claim that duck meat is delicious. In these cases, speakers faultlessly disagree \citep{lasersohn2005}, and in some cases there can never be a factual decision of what is true. In the case of Hamas, if a speaker says that October 7, 2023 was an act of resistance, then they do so by ignoring the rational declarative component that involves knowledge of acts of terror on a large scale. The human can do that by committing to irrational belief--- for instance that the facts didn't happen--- as is quite typically the case with conspiracy theories, where the speaker suppresses the declarative evidential component and works with the subjective component alone.

\smallskip

The  evidence for the relevance and importance of the subjective component emerges clearly in social networks interactions. Due to the lack of geographical co-location of the speakers and the impossibility to witness and acquire direct evidence for facts, reports. Often, what is dubbed `reported' evidence becomes of primary importance. Reported evidence comes as an hyperlink or the mention of a source. 
In a recent  study that correlates the use of speech acts and assertion in particular with types of sources of information \cite{boscaroetal2024phronesis} find that bare assertions overwhelmingly correlate with reported evidence. \cite{haan1999} explicitly mentions that indirect evidence correlates with a low degree of believability by a rational speaker, in comparison with direct evidence.

 Believability and trustworthiness are mental states that proceed from the evaluation of exogenous evidence in the light of endogenous beliefs and preferences. 
One and the same reported source can be trustful or not for two different speakers, depending on their background beliefs and expectations.

\cite{boscaroetal2024phronesis} define a measure for trustworthiness based on endogenous and exogenous evidence. Let $C$ be the exogenous content and $\mathcal{B}$  the set of subjective beliefs and preferences, i.e. the endogenous component Speakers assigns a measure of likelihood to a certain content $p$ (the content of their assertion), based on these two  sources: 

\ex. $\mu (p \mid  C \cap \mathcal{B})$ (Boscaro, Giannakidou and Mari 2024) 

$\mu$ is thus the measure of the veridicality judgment. If the speaker considers the content $C$ reliable, the measure assigned to $p$ will be 1 or close to one. This value will be lower otherwise. 

Given all the above, let us now conisder ChatGPT. ChatGPT certainly works on the basis of textual information, and, on the basis of this set it responds, provides information, calculates and produces informative sentences. ChatGPT+ certainly has content $C$. What ChatGPT crucially lacks if $\mathcal{B}$, that is to say the endogenous perspective, personal preferences, emotive content, that could allow it to form a veridicality judgment human style. 

But, worse, ChatGPT lacks the ability to form a veridicality judgment because it lacks the notion of truth.  ChatPGT can be taught to extract patterns in the data, it can also be rational in requiring system internal consistency, but it lacks referentiality to the world which is the proof process for truth. In order to be able to distinguish and know what is true, Cantwell Smith argues, the AI system would need to be `existentially committed', as the human mind is, by being directed to reality. This also means that ChatGPT would have to be able to distinguish between objects and their representations--- as well as between the actual, the possible, the necessary and the impossible, as we illustrated earlier with modal sentences. In order to have such representations one has to have internal states that are formed by empirically abstracting over what is the case in the world. Such abstraction is self-driven in humans, manifested clearly in early lexical categorizations in children. There is absolutely no evidence that a process of even remotely forming representations is true of ChatGPT. 

In the spirit of Aristotle, Cantwell Smith emphasizes that not only must AI be able to make factual distinctions but it must also be able to care about them;  and that would require both orientation to the world plus a nexus of normative and ethical norms and commitments. Without authentic engagement with the world, AI will not be able to exercise any kind of judgment judgment, which Smith defines as “dispassionate deliberative thought, grounded in ethical commitment and responsible action, appropriate to the situation in which it is deployed” (p. xv). In other words, ChatGPT lacks \textit{phronesis} (as defined in Aristotle's chapter 6, Nichomachean Ethics, as ethical decision making specific to particular problems and action, i,e. a form of instrumental and moral rationality, as \cite{ober2022greeks} calls it.

\section{Conclusions: truth, intention, deception}

In conclusion, ChatGPT does not have a mind in the sense of human mind. It lacks, as we said, \textit{logos}--- the ability to form internal representations that enable rational and moral judgment. It can be fed with credences and expectations, it can also be taught  to reproduce these including distinguishing contradictory beliefs; but ChatGPT cannot \textit{enhance} or self-drive this process, it can only reproduce it with reference to the data itself alone. It can reproduce behavior without featuring the internal self-driven \textit{causal} system that enables both veridical and moral judgment, which are at the core of our human intelligence. 

As we reach the end of our commentary, we want to point out two more things. Firstly, ChatGPT cannot deceive or mislead. We do not attempt here a fine-grained analysis of the distinction between deceiving and misleading or lying. They all involve the speaker knowing the truth but intentionally deciding to violate the Veridicality Principle (say what you know or believe to be true). These non-cooperative assertions require the intent to deceive and they presuppose the truth judgment, only to violate it.  Trivially, ChatGPT cannot lie because, lacking referentiality to the world, it does not `know' what truth is, as we established in the previous section. The only concept of truth it can have would be "truth is  what most commonly occurs in the data", something quite different from Aristotle's definition. But a statistically significant recurrent content is not necessarily true-- while it would be seen as true if it is consistent with other patterns expected by the algorithm.

ChatGPT can indeed mislead and deceive, we will argue, but not intentionally. It can mislead if it is fed with ambiguous, incomplete and not reliable information. It can also mislead in virtue of the fallacies of the algorithms that it relies upon. However, it cannot intentionally decide to mislead and deceive. The intentionality to deceive requires, to go back to our jargon from the previous section,  evaluating the trustworthiness of content $C$ with $\mathcal{B}$, and deciding to present $C$ as reliable in spite of a contrary assessment of $C$ through $\mathcal{B}$, and through checking with what is actually the case--- all moves that are simply against ChatGPT's nature.

\bibliographystyle{chicago}
\bibliography{intellecticabib}

\end{document}